\documentclass[conference]{IEEEtran}
\IEEEoverridecommandlockouts
\usepackage{cite}
\usepackage{amsmath,amssymb,amsfonts}
\usepackage{algorithmic}
\usepackage{graphicx}
\usepackage{textcomp}
\usepackage{xcolor}
\usepackage{tabularx}
\def\BibTeX{{\rm B\kern-.05em{\sc i\kern-.025em b}\kern-.08em
    T\kern-.1667em\lower.7ex\hbox{E}\kern-.125emX}}
\begin{document}

\title{Physics-informed Neural Networks approach to solve the Blasius function\\
}

\author{\IEEEauthorblockN{Greeshma Krishna}
\IEEEauthorblockA{\textit{Department of Mathematics} \\
\textit{Amrita Vishwa Vidyapeetham}\\
Amritapuri, India \\
greeshmakrishna@am.students.amrita.edu}
\and
\IEEEauthorblockN{Malavika S Nair}
\IEEEauthorblockA{\textit{Department of Mathematics} \\
\textit{Amrita Vishwa Vidyapeetham}\\
Amritapuri, India \\
malavikasnair@am.students.amrita.edu}
\and
\IEEEauthorblockN{Pramod P Nair}
\IEEEauthorblockA{\textit{Department of Mathematics} \\
\textit{Amrita Vishwa Vidyapeetham}\\
Amritapuri, India \\
pramodpn@am.amrita.edu}
\and
\IEEEauthorblockN{Anil Lal S}
\IEEEauthorblockA{\textit{Department of Mechanical Engineering} \\
\textit{Amrita Vishwa Vidyapeetham}\\
Amritapuri, India \\
anillals@am.amrita.edu}
}

\maketitle

\begin{abstract}
Deep learning techniques with neural networks have been used effectively in computational fluid dynamics (CFD) to obtain solutions to nonlinear differential equations. This paper presents a physics-informed neural network (PINN) approach to solve the Blasius function. This method eliminates the process of changing the non-linear differential equation to an initial value problem. Also, it tackles the convergence issue arising in the conventional series solution. It is seen that this method produces results that are at par with the numerical and conventional methods. The solution is extended to the negative axis to show that PINNs capture the singularity of the function at $\eta=-5.69$.
\end{abstract}


\section{Introduction}
Machine learning helps learn patterns from the data to make predictions. The three basic requirements for machine learning are data, theory, and hardware to overcome computational difficulties using better GPUs and new approaches. Computational science is an essential tool that we can use to incorporate physical invariances into learning. For example, the laws that govern the conservation of momentum, mass, and energy. To quote  Dr.Tinsley Oden - “Computational Science can analyze past events and look into the future. It can explore the effects of thousands of scenarios instead of actual experiments and be used to study events beyond the reach of expanding the boundaries of experimental science”. Deep learning can be quite useful in the real world, recognising various types of cancer like skin cancer  \cite{Mannava2022}, lung cancer \cite{Dev2019} and many more by just passing the necessary images. Also has various applications in agriculture \cite{bhanu2020machine}.

Neural networks(NNs) are the most used Machine Learning technique. It offers a set of robust tools for solving variously supervised and unsupervised problems in pattern recognition, data processing, and non-linear control, which can be regarded as complementary to the conventional approaches \cite{Bishop1994}. It can be considered a functional approximator as it maps a set of input variables to a set of output variables. A set of parameters called weights manages this mapping. These weights are updated to train the network that fetches the required output. A polynomial can be viewed as a function that transforms a single input variable into a single output variable. The coefficients in the polynomial are comparable to the weights in a neural network. Determining these coefficients helps in evaluating the solution.

\subsection{Physics Informed Neural Networks}

Physics-informed neural networks (PINNs) play a huge role where fewer data and the system's conventional physics are known to get the solutions to a differential equation. NN can approximate the error to ground truth function, generalize to unseen data and train the model. The contribution of PINNs is to obtain a neural network that knows about the physics hidden behind the equation and thus efficiently solves differential equations. The earlier numerical methods for solving them are finite difference, finite elements, spectral elements, and finite volume. When working with PINNs, there is data efficiency from a machine learning perspective since it is regularized heavily with physics. The required derivatives can be calculated using automatic differentiation at the end of the network. It is possible to train a nonlinear neural network using this approach and its layers of hidden nodes \cite{Nair2011}. It then finds a loss function corresponding to the differential equation and boundary conditions.

\subsection {Boundary Layer Theory}
The boundary layer of a flowing fluid is a thin layer near a solid surface, and the flow near the solid surface is known as the boundary layer flows. Ludwig Prandtl is credited for developing the boundary layer theory. In 1904, he published a paper titled "On the Motion of a Fluid with Very Small Viscosity" \cite{Prandtl1904}. He laid out the mathematical foundation for flows and condensed the two-dimensional Navier-Stokes equations (NSE) into the boundary layer equations. This publication made understanding fluid motion physics possible, which is regarded as the beginning of contemporary fluid mechanics \cite{LalS2021}. The primary disadvantage of using a CFD solver for turbomachinery optimization is the amount of time needed to complete the numerous computationally demanding simulations \cite{Abraham2022}.

The introduction of the similarity variables and the transformation of the PDEs into nonlinear ODEs in one coordinate allowed the successful resolution of the problem. Still, specific methods were needed to handle the unbounded boundary conditions. The Blasius benchmark problem has been resolved \cite{Mutuk2020} using the trial function method put out by Lagaris \cite{Lagaris1998} or a hybrid approach \cite{Ahmad2014}. The solution function is seen to have a singularity on the negative real axis at approximately -5.69. The Blasius equation is a single equation that solely models the viscous boundary layer and is one of CFD's fundamental models.

In this paper, we propose solving the Blasius equation using PINNs and comparing the solution we obtain with the best-known solutions available in the literature. We shall also extend our solution to the negative real axis to locate the singularity existing for the function. In the next section of the paper, we shall present a brief literature review on the Blasius equation and PINNs. In Section III, we present the methodology that has been adopted to solve the Blasius Equation. Section IV discusses the results and analysis of the proposed algorithm. The conclusions are discussed in the last section.

\section{Literature review}
\subsection{Physics Informed Neural Networks}
Neural networks can be used to attain solutions to ODEs and PDEs by reducing them to an optimization problem instead of numerically solving the equations. The application of NNs in fluid mechanics began in the 1990s. Raissi and his team \cite{Raissi2019, Raissi2017} developed a technique called Physics informed neural network (PINN) where the loss function defined in the corresponding NN is extracted from the physics behind the PDE and related equations. In earlier years, Computational fluid dynamics (CFD) had been a great relief in numerically solving the compressible and incompressible Naïve Stokes equations. Minimizing the loss function is challenging in PINNs, as it can be highly complex. Nevertheless, PINNs have been proven to be more accurate than conventional CFD techniques with lesser computations \cite{Bararnia2022}.

Although numerical discretization of the Navier-Stokes equations (NSE) has made significant progress in simulating flow problems over the past 50 years, the present algorithms cannot solve governed by high-parametrized NSE. Additionally, it is expensive to solve inverse flow problems due to their complexity, expensive formulations, and need for new algorithms. PINNs are expanded to fractional PINNs (fPINNs) that explore their convergence methodically to solve space-time fractional advection-diffusion equations (fractional ADEs) \cite{Pang2019}. A hybrid method for building the residual in the loss function that utilizes both automated differentiation for the integer-order operators and numerical discretization for the fractional operators is a novel component of the fPINNs.

Wide varieties of PINNs have been found in the literature in recent times. can-PINNs \cite{Chiu2022} link derivative terms with nearby support points, which generally apply to Taylor series expansion-based numerical systems. Apart from demonstrating good dispersion and dissipation characteristics, they are highly trainable and require four to sixteen times fewer collocation points than original PINNs. Auxiliary PINNs (A-PINN) is a technique for solving forward and inverse problems of non-linear integrodifferential equations \cite{Yuan2022}. ViscoelasticNet is another PINN framework for stress discovery and model selection \cite{Thakur2022}. PINNs can also be used to solve Reynolds-averaged Navier-Stokes equations \cite{Eivazi2022}, full waveform seismic inversions in 2D acoustic media, and wave propagation as it seamlessly handles boundary conditions and physical constraints \cite{RashtBehesht2022}. In addition to addressing ill-posed problems beyond the scope of conventional computing techniques, PINNs can also close the discrepancy between computational and experimental heat transfer \cite{Cai2021}.

\subsection{Blasius Equation}

Blasius equation is a third order non-linear ordinary differential equation of the form $f''' + \frac{1}{2} ff'' = 0$ with the boundary conditions $f(0)=0$, $f'(0)=0$, $f'(\infty)=1$. It governs the boundary layer flow over a semi-infinite flat plate. Suppose that the $u-$velocity, the velocity parallel to the surface, is much greater than the $v-$velocity, perpendicular to the surface, and the changes in the perpendicular direction to the surface are much greater than changes parallel to the surface. The boundary layer equations consists of conservation of mass (\ref{mass}), conservation of x-momentum (\ref{xmomentum}), and conservation of y-momentum. In a flat plate boundary layer, the pressure gradient term appearing in the x-momentum equation becomes zero (\ref{xmomentum}). This leads to the hydrodynamic solution for the flat plate boundary layer in a laminar flow called the Blasius solution. 
\setcounter{equation}{0}
\numberwithin{equation}{section}
\begin{equation}
\begin{gathered}\label{mass}
\frac{\partial u}{\partial x} + \frac{\partial v}{\partial y} = 0 
\end{gathered}
\end{equation}
\begin{equation}
\begin{gathered}\label{xmomentum}
u\frac{\partial u}{\partial x} + v\frac{\partial u}{\partial y} = \nu \frac{\partial^2 u}{\partial y^2}
\end{gathered}
\end{equation}

Blasius' analysis focus on the laminar boundary layer forming on a flat plate. The main aspect of the work is the transformation of the PDE for a flat plate boundary layer with zero pressure gradient into a single ordinary differential equation(ODE) by considering the velocity components that satisfy equation \ref{mass}.
\begin{equation}
\begin{gathered}\label{eq1}
u \equiv \frac{\partial \psi}{\partial y} \ \ \ \  v \equiv - \frac{\partial \psi}{\partial x}
\end{gathered}
\end{equation}
The stream function $\psi= U_\infty \sqrt{\frac{\nu x}{U_\infty }} f(\eta)$, is directly proportional to the function $f(\eta)$ called the Blasius function. Here, $U_\infty$ is the free stream velocity, and $\eta$ is a transformed coordinate called the similarity parameter. Here, the velocity components are proportional to the first derivative of $f(\eta)$, and the second and third derivatives of $f$ are proportional to the first and second derivatives of velocity. Substituting these relations into the momentum equation (\ref{xmomentum}), the final form of the Blasius boundary layer equation $f''' + \frac{1}{2}ff'' = 0$, for a flat plate can be obtained. The first and second derivatives of $f(\eta)$ are given by $f' = \frac{u}{U_\infty}$ (Non-dimensional velocity profile) and $f''=\frac{1}{U_\infty}\sqrt{\frac{\nu x}{ U_\infty}} \frac{\partial u}{\partial y}$ (quantity related to shear stress). The boundary conditions are set considering the laminar flow on a flat plate, the no-slip condition, and free-stream velocity outside the boundary layer. Hence we have 
\begin{equation}
\begin{gathered}\label{boundary}
f(0)=0\ \ \ \ f'(0)=0 \ \ \ \   f'(\infty)=1
\end{gathered}
\end{equation}
Since the value of $f''$ at $\eta=0$ is unknown, it cannot be considered an initial value problem.

A power series solution to the boundary layer equation of flow across a flat plate was proposed by Blasius \cite{Blasius1907}. Schmidt and Beckmann \cite{Pohlhausen1930} and Ostrach \cite{Ostrach1952} conducted the most important work on the topic, conducting theoretical and experimental research on the free convection flow of air around a vertical flat plate under the influence of gravity.
Boyd solved the equation using an analytical series solution technique \cite{Boyd1999}. Nowadays, with the availability of computers, we can obtain a numerical solution to this equation and calculate it with a very high degree of accuracy. To solve the Blasius equation numerically, a possible method is to use the shooting algorithm to find what value would satisfy the boundary condition at $\eta=0$. The first step is to guess a value at the wall. Then solving the ODE along the non-dimensional coordinate until the first derivative of $f$ stops changing. This checks if the first derivative satisfies the given boundary condition; instead, adjust the guessed value to decrease or increase if the first derivative is higher or lower than one. The algorithm is repeated until the boundary condition is finally satisfied to reach the final solution. Howarth \cite{Howarth1938} found the solution using a numerical method that accurately predicted the value of $f''(0)$ to be 0.332.

This differential equation is a direct representation of the velocity profile inside the boundary layer. Once the solution is obtained, the velocity components can be calculated using these relations. Hence the boundary layer thickness can be calculated located at the point where the velocity is 99$\%$ of the free stream velocity. It was found that this occurs approximately at $\eta$ equal to 5. The displacement and momentum thickness can also be estimated using the solution. The wall shear stress based on the velocity gradient at the wall can be estimated, and the friction and drag coefficients can be calculated. Note that the expression is only accounting for one side of the plate. Comparing the exact solution with the approximation solution from the integral analysis is interesting. The relations are close to each other. Indeed the integral analysis is within 10$\%$ of the exact solution; unlike Blasius' solution, these values were obtained without complicated math. The Blasius solution provides a self-similar solution meaning that the solution is the same if the independent and dependent variables of the governing equations are appropriately scaled. This can be seen in comparing experimental results with the Blasius solution. Different Reynolds numbers ($Re$) provide the same profile when the variables on the two axes are appropriately scaled. 

The solution to the Blasius equation can be found in \cite{ anil2014, Milin2022}, where the accurate benchmark results of the Blasius boundary layer problem using a leaping Taylors series that converges for all real values. There have been different methods to solve the Blasius equation, such as the Topher transformation, which is executed using inverse transformation, \cite{white2006viscous, Cortell2005, fang2008new}. Runge-Kutta, incorporated with the shooting method, finds the solution numerically, while Adomian Decomposition Method \cite{Abbaoui1995} finds the solution analytically. 

\section{Proposed methodology}
\begin{figure}[ht]
  \includegraphics[width=\linewidth]{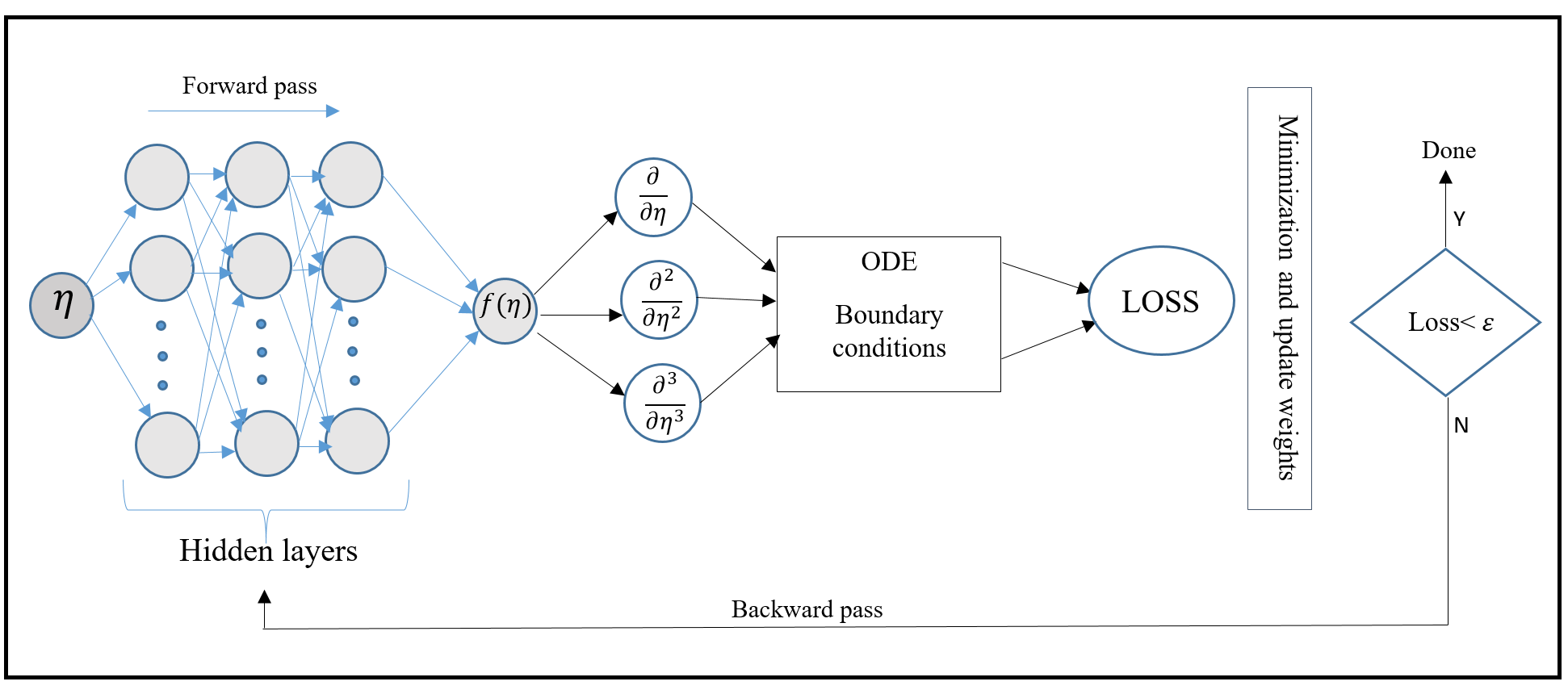}
  \caption{Workflow diagram to solve for $f(\eta)$ }
  \label{nn}
\end{figure}

The ODE representing the Blasius equation derived from the PDE is considered in the present work.  We follow a similar methodology proposed for the general form of generalized non-linear ODE by Raissi\cite{Raissi2019}. The workflow diagram of the present method is shown in Fig.\ref{nn}.

The input values $\eta$ are discrete and are equally distributed between the boundary points $\eta_0$ and $\eta_m$.  The beauty of PINNs lies in loss function as it incorporates the boundary conditions and the differential equation and thus includes the physics information. The loss function is the total loss from given ODE $L_o$, initial conditions $L_i$, and boundary $L_b$ conditions. 
\begin{equation}
\begin{gathered}
L=L_o+L_b+L_i
\end{gathered}
\end{equation}
\begin{equation}
\begin{gathered}
L_o= \sum_\eta{[\hat{f}'''+\frac{1}{2}\hat{f}\hat{f}'']^{2}}
\end{gathered}
\end{equation}
\begin{equation}
\begin{gathered}
L_i=[\hat{f}(\eta_0)-f(\eta_0)]^{2} + [\hat{f}'(\eta_0)-f'(\eta_0)]^{2}\\
 =  [\hat{f}(\eta_0)-0]^{2} + [\hat{f}'(\eta_0)-0]^{2}
\end{gathered}
\end{equation}
\begin{equation}
\begin{gathered}
L_b= [\hat{f}(\eta_m)- f(\eta_m)]^{2} =  [\hat{f}(\eta_m)-1]^{2}
\end{gathered}
\end{equation}
From the definition of the loss function, it is clear that the total loss will be zero if the function $f(\eta)$ is exact. Here we update the weights of the neural network in each iteration such that the loss function is minimized. Supplying the independent input values representing discrete spatial coordinates $\eta$ ranging from zero to $\eta_m$ into the neural network is sufficient to solve the ODE via PINNs. The neural network maps the input $\eta$ to $f(\eta)$, which is the estimated solution to the stream function. Unlike standard neural network techniques that approximate the value of $f(\eta)$ in a heuristic manner from sample output values, PINNs obtain a solution function that minimizes the loss function, which is a combination of the differential equation and boundary values. 

\begin{figure}[ht]
  \includegraphics[width=\linewidth]{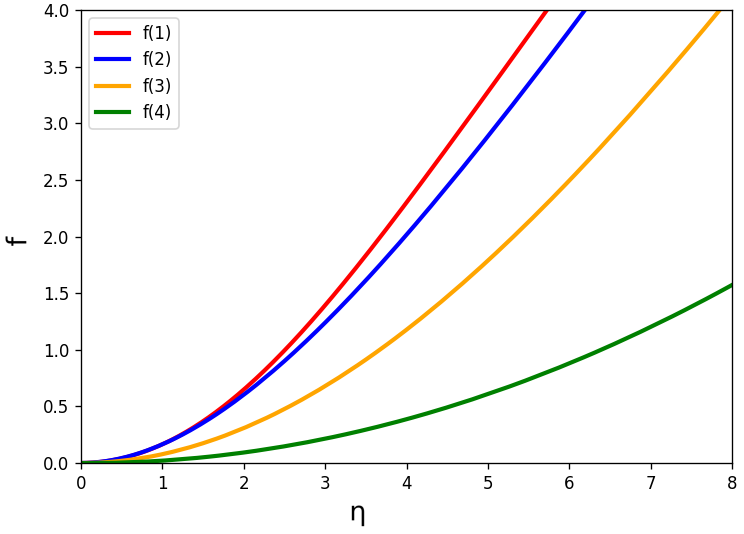}
  \caption{Solutions obtained for various widths and depths of NN}
  \label{compare}
\end{figure}

Several combinations of the number of hidden layers and the number of nodes in each layer were tested on a trial-and-error basis. We see that the results obtained are at par with that of \cite{ Milin2022} when initializing the neural network using two fully connected hidden layer structures and setting each hidden layer's width to 100 neurons. The learning rate set in this case is 0.96. Some of the relatively good results obtained for various combinations of hidden layers and nodes, along with the changes in the learning rate of the algorithms, are presented in table \ref{tab1}. The solution $f_i(\eta)$ for each of the cases in table \ref{tab1} is graphically presented in  Fig.\ref{compare}.

\begin{table}[ht]
\caption{Variations in $f''(0)$  and Loss Function}
\begin{center}
\begin{tabular}{ |c|c|c|c|c| } 
 \hline
Layers & Neurons &  f''(0) & Loss & Solution\\
\hline
2  & 100  & 0.33165 & 1.67x10$^{-6}$ & $f_1(\eta)$ \\
\hline
3  & 100  & 0.32154 & 1.10x10$^{-4}$ & $f_2(\eta)$ \\
\hline
4  & 90  & 0.16021 & 1.95x10$^{-3} $& $f_3(\eta)$ \\
\hline
5 & 90  & 0.0.5071 & 5.58x10$^{-3}$ & $f_4(\eta)$\\
\hline
\end{tabular}
\label{tab1}
\end{center}
\end{table}

In earlier methods proposed by Lagaris\cite{Lagaris1998}, all the boundary conditions had to be found before moving to find the solution. In this method, it is only required to set an appropriate finite value for the $\eta$ at $\infty$, which we assumed to be $\eta_m=8$ in our case. Although higher values could be set, we see from the literature and our results that $f'(\eta)$ takes the value very close to one when $\eta=5$. Hence the justification for taking the value of $\eta$ at $\infty$ as $\eta_m=8$. Another benefit of using PINNs is that we can considerably reduce the number of collocation points and still attain the same level of accuracy. In our method, we have considered a total of 100 equidistant collocation points of $\eta$ between 0 and 8. The network trains with this input data and finds the derivatives using the automatic differentiation tool to calculate the loss $L_o$.

The Adam optimizing algorithm is incorporated into the PINNs methodology to set the model's adaptive learning rates. It adds momentum as the estimate of the first-order moment of the gradient and includes bias corrections to the estimates of first and second-order moments. The second-order method to train the network was the Limited memory Broyden Fletcher Goldfarb Shanno  (L-BFGS) algorithm. The cost of memory has been decreased by avoiding the hessian approximation of BFGS and replacing it with an identity matrix. The loss function is optimized using ADAM and L-BFGS until it converges.

The method was extended to the negative axis setting $\eta_0$ to -5.69 and $\eta_m$ to 7. The value of the $f''(0)$ found from the previous program is incorporated into the existing loss function. Our purpose in doing this was to check if our method could capture the singularity of the solution function mentioned in the literature.

\section{Results and discussion}

The proposed PINNs method can reduce computational time in solving the Blasius equation over conventional CFD techniques as it reduces the burden of finding the value of $f''(0)$ and then solving the initial value problem. 
\begin{figure}[ht]
  \includegraphics[width=\linewidth]{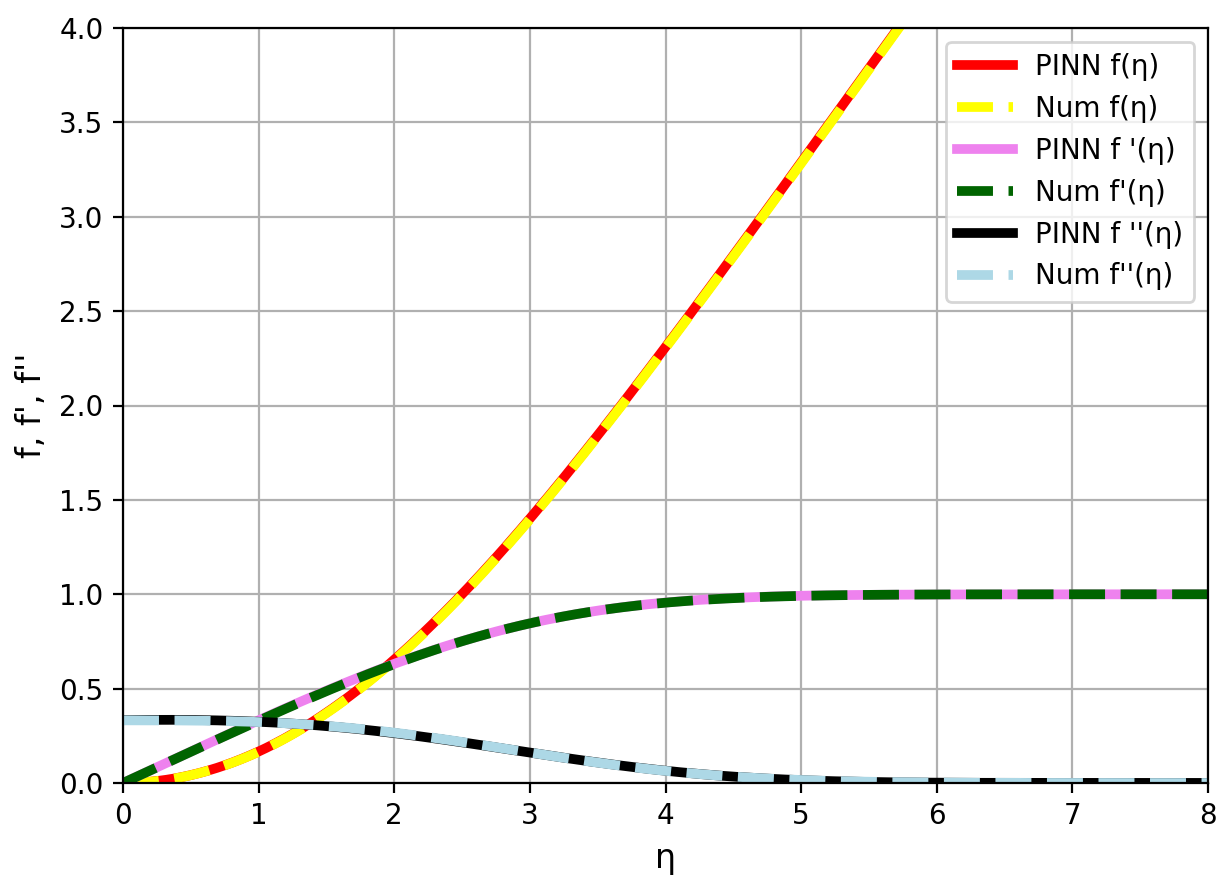}
  \caption{Estimated results $f(\eta), f'(\eta), f''(\eta)$ obtained from PINNs and those obtained by numerical methods}
  \label{Estimated results $f, f', f''$ with numerical solution}
\end{figure}
The neural network constructed with two hidden layers with 100 neurons each is considered for our results and discussions. The input of $\eta$ ranges from 0 to 8 ($\eta_m$). The learning rate for adam optimizer is assigned the value of 0.96. From Fig.\ref{Estimated results $f, f', f''$ with numerical solution}, it is clear that the results obtained from PINNS perfectly match with numerical techniques' results \cite{anil2014}. 

 In the present method, the value of $f''(\eta)$ at 0 is not considered for finding the solution. Further, once $f(\eta)$ is obtained, we can also find the functions corresponding to $f'(\eta)$ and $f''(\eta)$. The value of the second derivative at the wall is observed from the PINNs method to be 0.33165, which aligns approximately with those found by Howarth \cite{Howarth1938}. The loss generated from this method is 1.67x10$^{-6}$ that is of order $10^{-6}$ indicating the results to be accurate with sufficiently less error value. As proposed by Gaurav Pandey and Ambedkar Dukkipati \cite{pandey2014go}, increasing the width (number of neurons) and depth (number of hidden layers) of the network can be beneficial. But in our case, we see that the function $\hat{f}(\eta)$ tends to become more vulnerable to a high amount of bias by choosing a deeper network. Thus, the initial conditions must be chosen much more accurately for the loss function to converge to the global minimum. Further, fixing the width and depth of the network to 100 and 2, respectively, depicts accurate results; hence, the need for a more complex network was overridden. 

The velocity components are directly proportional to $f'(\eta)$. Thus, the proposed method's solution helps in mapping non-dimensional coordinate $\eta$ with $f'(\eta)$. This is the velocity profile inside the boundary layer. The values of velocity components can also be calculated from $f'(\eta)$ and $f''(\eta)$ From the graph, it is visually appealing that when $\eta$ approaches five, the $f'(\eta)$ goes to 1. Hence the boundary layer thickness is calculated where the velocity is 99$\%$ of free stream velocity. 
\begin{figure}[ht]
  \includegraphics[width=\linewidth]{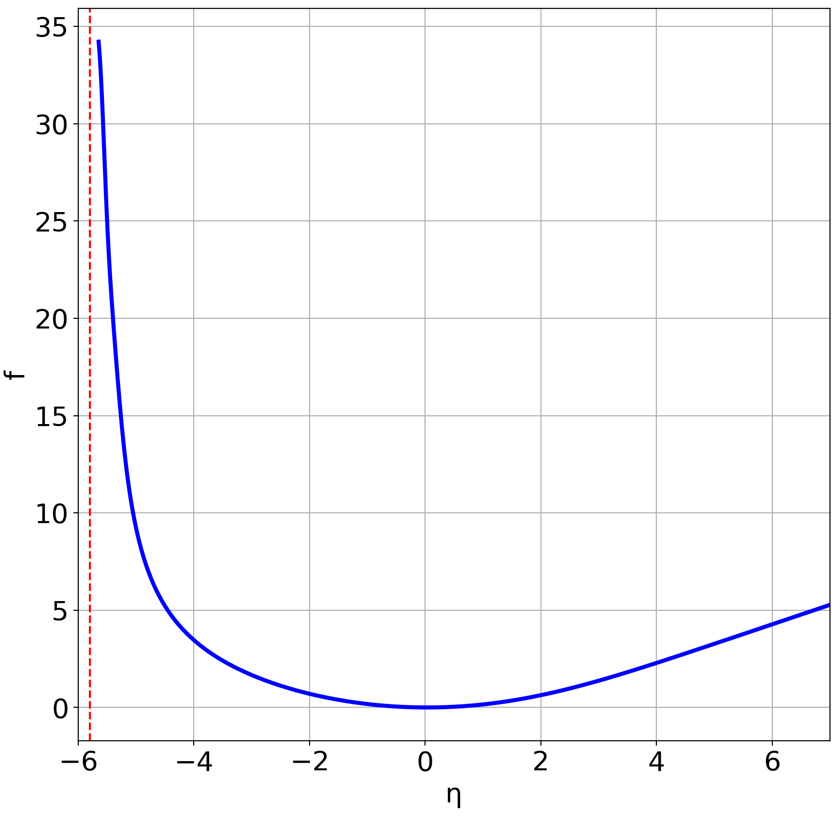}
  \caption{Graph of $f(\eta)$ on the negative $\eta-$axis}
  \label{negative}
\end{figure}

In 1999, Boyd \cite{Boyd1999} suggested that convergence of his power series method to solve the Blasius equation is limited by a singularity on the negative $\eta-$axis at $\eta=-5.6900380545$. Anil and Milan \cite{ Milin2022} considered a leaping Taylor's series solution for the Blasius equation to overcome this singularity. In the second part of our investigation, we extended the collocation points to the negative portion to check for the solution of the Blasius equation on the negative $\eta-$axis. The solution obtained from PINNs also shows that the function $f(\eta)$ increases rapidly near $\eta=-5.7$ signifying the presence of a singularity near the point. A graph of $f(\eta)$  including the negative $\eta-$axis is shown in Fig.\ref{negative}.

\section{Conclusion}
 PINNs use knowledge of the governing equation in deep learning and find a solution to the differential equation by minimizing a loss function, including the physics information.  The proposed methodology using PINNs is free from mesh generation, which is an integral part of conventional CFD techniques. Here, we have obtained the solutions to the Blasius equations, which agree with all the numerical methods in the literature. Further, the solution captures the singularity mentioned while solving the differential equation analytically. 
 
\bibliographystyle{IEEEtran}
\bibliography{bibfile}
\end{document}